\documentclass[runningheads]{llncs}
\usepackage[utf8]{inputenc}
\usepackage[english]{babel}

\usepackage{multirow}  

\usepackage{amsmath}
\usepackage{booktabs}
\usepackage{soul} 
\usepackage{xcolor}

\usepackage{graphics}
\usepackage{subfig}
\newcommand{\da}[1]{{#1}}
\newcommand{\fm}[1]{{#1}}

\newcommand{\B}{\bfseries}

\usepackage{graphicx}
\begin{document}
\title{Am I fit for this physical activity? Neural embedding of physical conditioning from inertial sensors}
%
%

 
\author{Davi Pedrosa de Aguiar \orcidID{0000-0003-2286-5841} \and\\
 Fabricio Murai \orcidID{0000-0003-4487-6381}}
 \authorrunning{D. Aguiar \and F. Murai}
 \titlerunning{Encoding Physical Conditioning for Heart Rate Estimation}
 \institute{Universidade Federal de Minas Gerais, Brazil\\ 
 \email{\{daviaguiar,murai\}@dcc.ufmg.br}
}
\maketitle              
\begin{abstract}
\small\baselineskip=9pt
  Inertial Measurement Unit (IMU) sensors are present in everyday devices such as smartphones and fitness watches. As a result, the array of health-related research and applications that tap onto this data has been growing, but little attention has been devoted to the prediction of an individual's heart rate (HR) from IMU data, when undergoing a physical activity. Would that be even possible? If so, this could be used to design personalized sets of aerobic exercises, for instance. In this work, we show that it is viable to obtain accurate HR predictions from IMU data using Recurrent Neural Networks, provided only access to HR and IMU data from a short-lived, previously executed activity. We propose a novel method for initializing an RNN's hidden state vectors, using a specialized network that attempts to extract an embedding of the physical conditioning (PCE) of a subject. We show that using a discriminator in the training phase to help the model learn whether two PCEs belong to the same individual further reduces the prediction error. We evaluate the proposed model when predicting the HR of 23 subjects performing a variety of physical activities from IMU data available in public datasets (PAMAP2, PPG-DaLiA). For comparison, we use as baselines the only model specifically proposed for this task and an adapted state-of-the-art model for Human Activity Recognition (HAR), a closely related task. Our method, PCE-LSTM, yields over 10\% lower mean absolute error. We demonstrate empirically that this error reduction is in part due to the use of the PCE. Last, we use the two datasets (PPG-DaLiA, WESAD) to show that PCE-LSTM can also be successfully applied when photoplethysmography (PPG) sensors are available, outperforming the state-of-the-art deep learning baselines by more than 30\%.

\end{abstract}
\section{Introduction}\label{sec:intro}
In the recent years there has been an ever increasing usage of sensor-equipped devices, such as smartphones, smartwatches and fitness watches. These sensors can be used to track user behavior and health-related measurements. Among the most common types are the Inertial Measurement Units (IMU), composed primarily of accelerometers and gyroscopes.The use of photoplethysmography (PPG) sensors to track heart rate (HR) is also becoming ubiquitous, especially in devices targeting fitness conscious consumers, such as the Apple Watch, FitBit and Samsung SimBand~\cite{reiss2019deep}.
%
To promote an effective fitness training, it is necessary to induce an optimal cardiovascular response, making it essential to model and predict individual HR responses~\cite{ludwig2018measurement}.

Several methods have been designed to predict the HR under the influence of physical activity. Most of them are based on differential equations~\cite{cheng2007de,hunt2016de}, Hammerstein and Wiener models \cite{mohammad2011hm_cycling} or Neural Networks. 
Among the latter, some predict many steps into the future \cite{zhang2018multihr_bayers,xiao2011hr_multistep}, and others use IMU signals as input
~\cite{yuchi2008heart,xiao2011hr_multistep} but only the model proposed in \cite{xiao2011hr_multistep} does both, to the best of our knowledge.

Despite the recent advances in neural networks, the problem of multi-step HR estimation from IMU sensor data remains little explored. 
In this paper, we investigate neural architectures that could be used for this task. 
To benchmark our model, in addition to using the model from \cite{xiao2011hr_multistep}, we adapt a network proposed for a closely related task called Human Activity Recognition (HAR), where the goal is to identify the activity being performed by a person (e.g., running, walking, swimming) given some sensor data (mostly, IMU) as input.  

Recurrent Neural Networks (RNNs) are widely used for multi-step predictions because they can carry information regarding the ``hidden'' state of a sequence through time as vectors. Although the performance of RNNs can be highly dependent on the initialization of these vectors, most often they are simply set to zero. Sometimes, the RNNs are set up so that the state vectors go through some iterations (washout period) before the first prediction can be returned. In the context of aerial vehicle dynamics, \cite{rnn_init_ds} proposes an approach to initialize these vectors by using features and labels from a previous interval.

In a similar spirit, based on the premise that different subjects will display different HR levels when performing the same exercises depending on their physical conditioning, we propose a novel neural architecture that attempts to encode that attribute given IMU and HR data collected from a previous, short-lived, activity performed by an individual. This attribute is extracted by a CNN network as a vector dubbed \textit{physical conditioning embedding} (PCE). The PCE is used, in turn, to set the initial values of the hidden and cell states of an LSTM responsible for outputting the HR predictions. Our proposed method, PCE-LSTM, differs from \cite{rnn_init_ds} in several ways, markedly, (i) it uses a discriminator to improve the ability of the network to capture the inherent characteristics of the subjects based on previous activities and (ii) it is able to improve predictions over windows of up to 2 hours. This corroborates the idea that PCE-LSTM is encoding physical conditioning, rather than information about specific physical activities.

Although PPG sensors for tracking HR are becoming more common, they are prone to measurement errors due to motion-related artifacts. In this case, data from other sensors can be used to correct the HR measurements. This task, called PPG-based HR estimation, has been explored in~\cite{salehizadeh2015,schack2017,reiss2019deep}. To show that the proposed architecture can also be used for this task, we adapt PCE-LSTM to incorporate the PPG signal as an additional input and show that it can also outperform the state-of-the-art in this task. 
In sum, our main contributions are: 
A \textbf{new model} to predict the HR from IMU-sensor signals, which outperforms state-of-the art baselines; An \textbf{ablation study} on the contribution of the PCE subnetwork to the performance of the model; A demonstration that PCE-LSTM  estabilishes the \textbf{new deep learning SOTA of  PPG-based HR estimation}.\footnote{We released all our code at \fm{\url{https://github.com/davipeag/HeartRateRegression}}.}

\paragraph{Outline.} Section~\ref{sec:methods} describes datasets and pre-processing used in this work. Section~\ref{sec:pce-lstm} details the proposed method, PCE-LSTM, and the key hypothesis we investigate. Section~\ref{sec:baselines} discusses the reference methods for the HR estimation task. Section~\ref{sec:evaluation} presents the evaluation results. The related work is reviewed in Section~\ref{sec:related}. Last, Section~\ref{sec:impact} discusses the significance and impact of this work.

\section{Methodology}\label{sec:methods}

This work addresses the problem of predicting the heart rate $H_t$ of an individual at time $t=1,\ldots$, given IMU sensor data gathered up to time $t$ and HR values from an initial, short lived period,  $H_1 \ldots H_{I}$, with $I < t$. We refer to this task as \textbf{IMU-based multi-step HR estimation}. For this study, we use the PAMAP2 and PPG-DaLiA datasets (Section~\ref{sec:datasets}), which are among the very few publicly available sets containing both IMU and HR signals from individuals performing a variety of activities. PPG-DaLiA also has data from PPG sensors, which is disregarded in this first task.

We address the case where PPG sensors are available (in addition to IMU) as a secondary prediction task, referred as \textbf{PPG-based multi-step HR estimation}. Since PPG data are HR measurements that can be perturbed by movement, this task consists of correcting such measurements based on IMU data. For this task, we use the (complete) PPG-DaLiA~\cite{reiss2019deep} and WESAD~\cite{WESAD} datasets. We show that it is possible to adapt the architecture proposed for the first to the second task with minor changes. 
Below we describe the datasets and the pre-processing techniques used throughout this work.

\subsection{Datasets}\label{sec:datasets}

\textbf{The PAMAP2 Dataset \cite{reiss2012dataset}} consists of data from 40 sensors (accelerometers, gyroscopes, magnetometers, thermometers and HR sensor) sampled at 100Hz of 9 subjects performing 18 different activities (e.g., rope jumping, running, sitting). There is a single time series of sensor signals per subject, each performing a sequence of activities. Later on we explain that the time series for 1 of 9 the individuals is too short for training the models.

\textbf{The PPG-DaLiA Dataset \cite{reiss2019deep}} is composed of signals from two devices, a chest-worn device which provides accelerometer and ECG data; and a wrist-worn device measuring the PPG and triaxial acceleration, sampled at 32 Hz. It also includes HR series computed from the ECG signals. This dataset contains a contiguous time series of sensor signals from 15 individuals performing 8 activities. We dub DaLiA,  the subset of this dataset which does not include the PPG signals (to use in the IMU-based multi-step HR estimation task). 

\textbf{The WESAD Dataset \cite{WESAD}} consists of data from 15 subjects wearing the same sensors available in PPG-DaLiA, but the individuals remain seated/standing during the whole study while going through different affective states (neutral, stress, amusement). Unlike PPG-DaLiA, WESAD does not provide precomputed HR series, therefore we used the heartpy library~\cite{van2019analysing} to extract HR measurements from the ECG signals. Although subjects are indexed up to number 17, the dataset does not include subject identifiers 1 and 12.

In total, we use data from 23 (resp.\ 30) subjects for IMU-based (resp.\ PPG-based) multi-step HR estimation task.

\subsection{Pre-processing}
\label{sec:preprocessing}

\textbf{Basic Preprocessing.} We upsample the HR signal using linear interpolation in all datasets to make its sampling rate consistent with the other signals. Only PAMAP2 contains a few missing data points, which we handle by local averaging the data around the missing point using a 0.4s window~\cite{steven2018feature}. To make the use of the PAMAP2 and PPG-DaLiA datasets more consistent, we use only the accelerometer signals of the chest and wrist, and downsampled the signals to 32Hz. All signals $\mathbf{s}$ are z-normalized, i.e., $\hat{\mathbf{s}} = (s - \mu(\mathbf{s}))/\sigma(\mathbf{s})$, where $\mu$ stands for the mean operator and $\sigma$ for the standard deviation operator.

\begin{figure}[t]
\centering
\includegraphics[width=0.5\columnwidth]{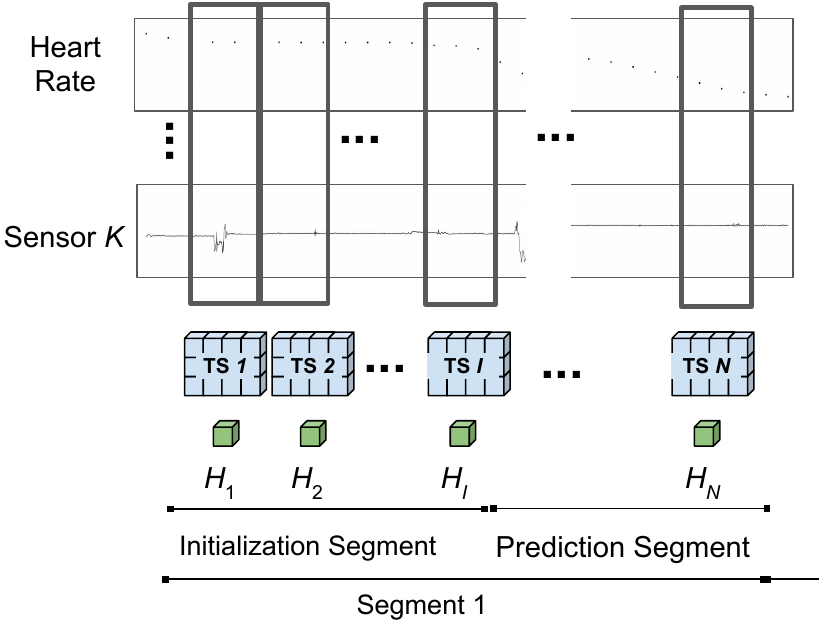}
\caption{Time Snippet Representation}
\label{fig:ts}
\end{figure}

\textbf{Time Snippet Discretization.} Like most works based on this data, we discretize the time series signals into \textbf{time snippets} (TS), i.e., partially overlapping windows of fixed duration $\tau_{TS}$ and overlap ratio $r_{TS}$ (task-dependent). Figure \ref{fig:ts} illustrates this procedure for the case when $r_{TS}=0$. Each time snippet $TS_t$ is a matrix where each row represents a sensor. Time snippets determine the granularity of the predictions. Accordingly, we define the average HR ${H}_{t}$ for each time snippet $\textrm{TS}_t$ as the response to be predicted. 

\textbf{Time Series Segmentation.} In order to create a fixed-length training set, we segment the time series of each individual in a sequence of $N$ contiguous time snippets of fixed duration. 
Each segment is partitioned into two smaller segments. The first, called \textbf{Initialization Segment}, contains the IMU and HR signals for the first $I=12$ time snippets, and can be used by a neural network (NN) to encode a ``state'' specific to that time series. The second, called \textbf{Prediction Segment}, contains the IMU, but does not contain the HR, as it is used by a (possibly different) NN to output predictions for each time snippet. Figure~\ref{fig:ts} illustrates the subdivision of a segment, in the case where $r_{TS}=0$. Note that the NN used for processing the first segment can also be used for processing the second segment by replacing the HR signal in the latter by zeroes.

\section{The Physical Conditional Embedding LSTM model} \label{sec:pce-lstm}

Here we describe PCE-LSTM, our proposed neural network architecture for HR prediction. PCE-LSTM is composed of convolutional and LSTM layers, similarly to the state-of-the-art techniques for the closely related task of HAR. The main novelty of this model is discussed below.

Recurrent Neural Networks (RNNs) have been especially designed to work with time series data. They are composed by cells with shared parameters, which process units of the input sequentially, using one or more vectors to carry state information through time. In particular, the unidirectional LSTM (long short-term memory) cell uses two vectors -- the hidden state and the cell state -- which are received from the previous iteration, updated based on the input for that time and passed onto the next iteration. The first iteration, however, receives these vectors as they were initialized, typically as zero vectors. The implicit assumption is that the network will gradually be able to encode the correct state from the inputs as the vectors are passed through the cells.

For HR prediction, we argue that if some data on the relationship between the input and output signals is available prior to the prediction, it can be beneficial to use a specialized network to encode this relationship as the RNN initial state. This initial state should contain information about physical conditioning: a more fit individual is able to sustain similar movement levels with smaller increase in HR.
    Thus, the main hypothesis investigated here can be stated as

\begin{changemargin}{}{}     
    \emph{\textbf{Hypothesis:} It is possible to encode information about physical conditioning from an individual's sensor data as a vector and use it as the initial state of a RNN to improve HR predictions.}
\end{changemargin}
This is the rationale behind the main difference between our approach and similar ones, namely the initialization of the RNN hidden vectors using a specialized network, which we call the Physical Conditioning Encoder.

\begin{figure}[t]
\centering
\includegraphics[width=.6\columnwidth]{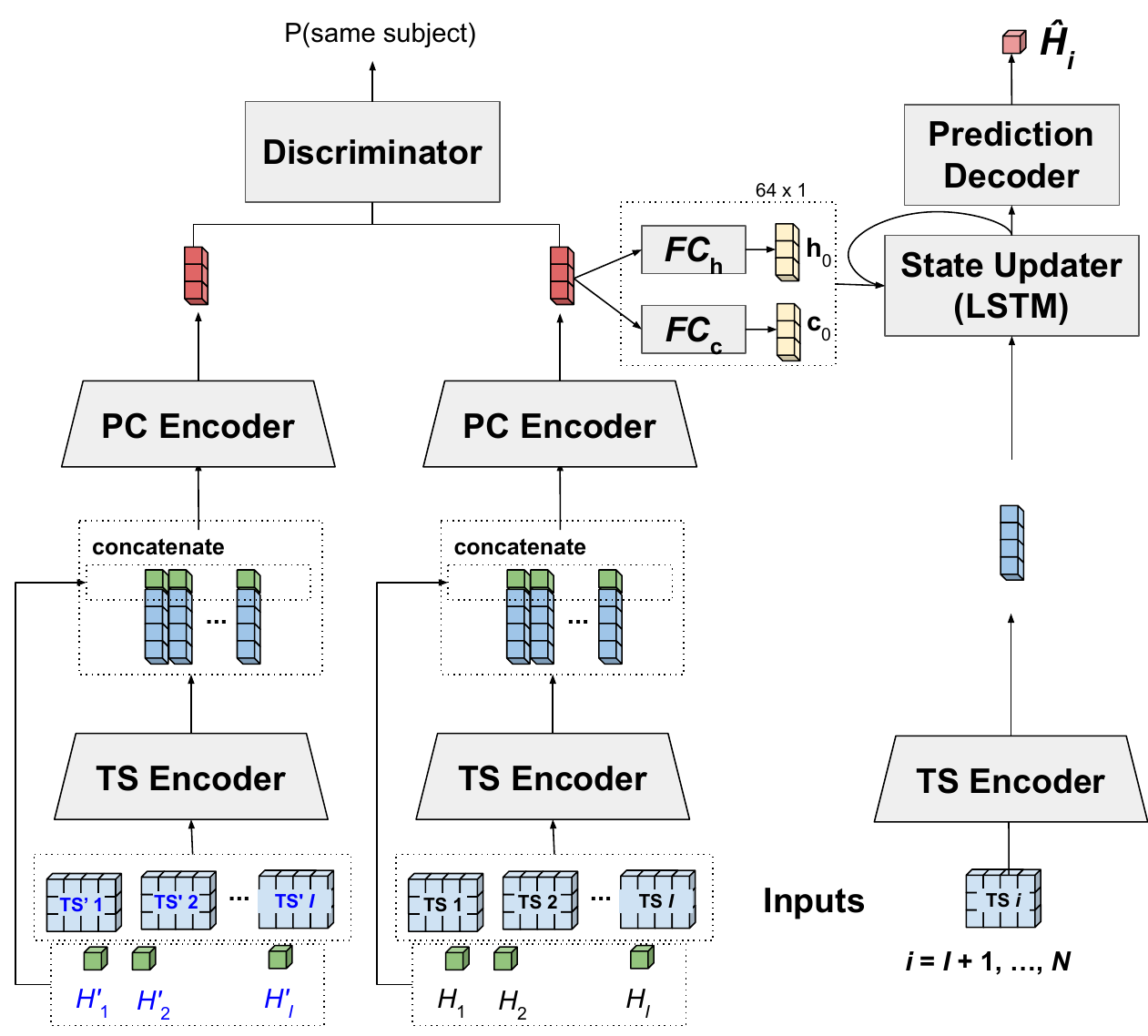}
\caption{PCE-LSTM (proposed architecture).}
\label{fig:ourConvLSTM}
\end{figure}

Figure~\ref{fig:ourConvLSTM} shows the high level structure of our architecture, which is made of five components: 
the \textbf{Time Snippet Encoder} (TS Encoder), a convolutional network which encodes the 2-dimensional TS into a vector; the \textbf{Physical Conditioning Encoder} (PC Encoder), a convolutional network which extracts, from signals of the Initialization Segment (including HR), a Physical Conditioning Embedding (PCE) used for initializing the LSTM's hidden vectors; the \textbf{Discriminator}, used to force the PCE of a given subject to be similar across time segments; the \textbf{State Updater}, a LSTM which maintains and updates the subject's state, encoded in its hidden vectors; and the \textbf{Prediction Decoder}, a Fully-Connected network which decodes the HR prediction from the state tracked by the State Updater. \da{We emphasize that instances of the same module shares the same weights.} 
%
Each component is described in detail below.

\textbf{TS Encoder:} extracts features from one time snippet. We reckon the role of each sensor in the description of the intensity of an activity as being equivalent to role of each RGB channel in the description of a picture. Hence, it is reasonable to combine them early in the NN. To do so, we use 1D convolutions along the time dimension, with the sensors stacked along the channel dimension. This approach differs from the most well known architectures in the literature~\cite{reiss2019deep,moya2018convolutional,ordonez2016convlstm} which keep the transformations on each sensor's signals separate during the convolutional section of their architectures by the use convolutional kernels spanning a single sensor, only combining them later on. 
    
    The TS Encoder comprises multiple layers, each made of a 1D convolution followed by a Leaky Rectified Linear Unit activation and a dropout layer of rate TSE$_{\text{dropout}}$. After each layer, the tensor length $\ell$ is reduced to $\lfloor \ell / 2 \rfloor$ by filters of size 3 and stride 2 (when $\ell$ is even, we use $\textrm{padding}=1$), except when the the tensor length is 2, in which case we use a filter of size 2 without padding. All layers have TSE$_{F}$ filters, except the last layer, which has TSE$_\text{out}$ filters. The number of layers is $\textrm{TSE}_{N}=\lfloor\log_{2}(\text{TS}_L)\rfloor$, where TS$_L$  is the length of the time dimension of the TS, so as to transform the size of the time dimension to one.
    
    The vectors extracted from each time snippet of the Initialization Segment are concatenated along the time dimension, and the HR of each TS is concatenated along the feature dimension before being passed onto the next component.

\textbf{PC Encoder:} takes the vectors concatenated in the previous step and extracts a \textbf{physical conditioning embedding} (PCE). It is a multi-layer convolutional network that transforms the 2D-input ( $(\text{TSE}_{\text{out}}+1) \times I$) into a single vector of length PCE$_\text{out}$. Its convolutional architecture is designed with the same principles as the TS Encoder. Figure~\ref{fig:ourConvLSTM} shows the PC Encoder in detail.

From the PCE, the LSTM's  hidden state and cell state vectors are computed using a single linear layer each, represented in Figure~\ref{fig:ourConvLSTM} by FC$_h$ and FC$_c$.

\textbf{State Updater:} is a standard LSTM with both state vectors (cell and hidden state's) of size LSTM$_H$, and input of size TSE$_{\text{out}}$.  These state vectors are initialized by the PC Encoder using only signals from the Initialization Segment. The LSTM is then fed with the deep attributes extracted by the TS Encoder from each time snippet.

 \textbf{Prediction Decoder:} takes the hidden state representation for each time snippet from the State Updater and computes the prediction from these representations. It is made of three fully connected layers, where the first two layers have 32 neurons, each followed by a ReLU activation function, and the last layer have LSTM$_H$ neurons, without activation function, outputting the predicted HR. We use the mean absolute error ($\ell_1$ loss) as the cost function $L_\text{HR}$ associated with this output because using the $\ell_2$ loss hampers training as larger differences between the predicted and actual HR have an out-sized impact on the loss, according to our preliminary experiments.

\textbf{Discriminator:} Given the benefits of multi-task learning we propose the simultaneous training of the model on a constructed task in which the PCE is used to distinguish between individuals. Assuming a subject's physical conditioning as constant in the short term, we reason that employing a network to discriminate whether two PCEs belong to same person will foster better embeddings, when trained jointly with PCE-LSTM\footnote{We avoid overfitting by using disjoint sets of subjects for training and validation.}. To train the Discriminator, for each segment in the training set, we sample another segment from the same individual with probability 0.50 and from a different individual with probability 0.50. For each pair, we measure the cross entropy as the loss function ($L_D$). We set the weight of the discrimination loss to 10\% of the total loss.\footnote{Alternatively, losses' weights can be set by hyperparameter tuning, but since we use a single subject for validation, we fixed the weights to $(0.9,0.1)$ to avoid overfitting to the validation subject.} Hence, the total loss $L_\text{total}$ is given by $L_\text{total} = 0.9 L_\text{HR} + 0.1 L_D $. This component is the head of a Siamese network and is completely optional, being used exclusively to improve the performance of the model.
The Discriminator's chosen architecture is comprised of 5 fully connected linear layers, each with 64 neurons and followed by ReLU activation function and a dropout rate of 0.15, except for the last layer, which uses a sigmoid activation function. This network receives two PCEs (concatenated) and outputs a probability. 

\subsection{Adaptations for PPG-based HR Estimation}

\begin{figure}[t]
    \centering
    \includegraphics[angle=0,width=1\columnwidth]{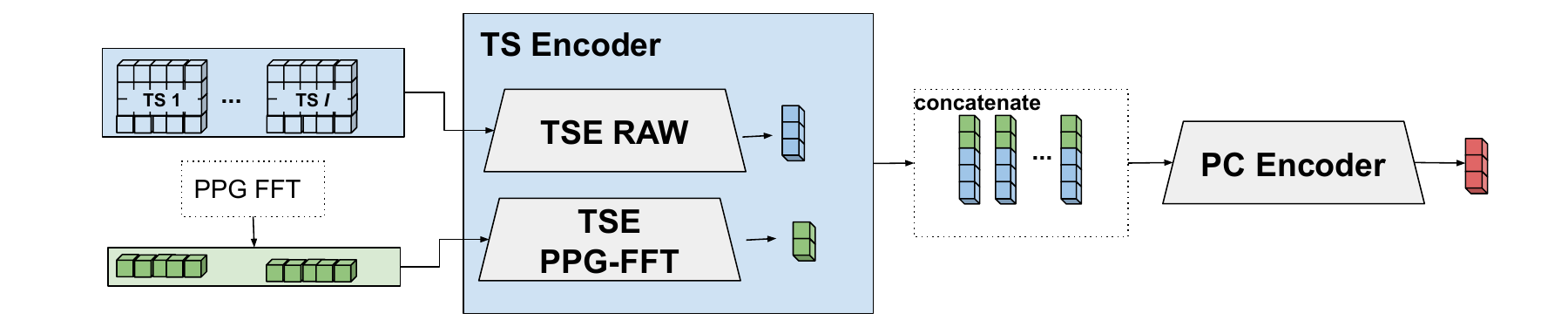}
    \caption{PPG-adapted PCE}
    \label{fig:PPG-PCE}
\end{figure}

The task of PPG-based HR estimation has 3 main distinctions from the IMU-only HR estimation and hence requires a few adaptations to our method.
First, the PPG signal is best represented in the frequency domain; second, the PPG signal is more important than the other signals since it is a rough estimate of the HR; third, for PPG-based HR estimation, a ground truth HR is not usually available. To deal with these differences, we modified the TS Encoder sub-network to comprise two TS Encoder architectures, one for the raw PPG and IMU signals ($\text{TSE}_{\text{raw}}$) and another for the Fast Fourier the Transformed PPG signal ($\text{TSE}_{\text{PPG-FFT}}$) . These outputs are concatenated and returned by the new sub-network. The $\text{TSE}_{\text{PPG-FFT}}$ is smaller, with $[\text{TSE}_{\text{PPG-FFT}}]_{\text{out}} = 12$. We also modified the PC Encoder slightly, so that it only receives as input the concatenated outputs of the TSE (without the HR). Figure \ref{fig:PPG-PCE} illustrates the changes to the PCE-LSTM architecture for the secondary task.

\section{Reference methods} \label{sec:baselines}
\label{sec:benchmark}

To the best of our knowledge, the task of predicting  multiple steps of HR given IMU signals has only been attempted by \cite{xiao2011hr_multistep}, which used in their experiments a dataset not publicly available.  In addition to using their model as a baseline, we also adapt a model \cite{ordonez2016convlstm} designed for similar tasks (HAR) in order to benchmark our method. 
In this section we describe each of these models, the task they were designed to address and the minor changes required to adapt the models for the task at hand. More recent architectures for the HAR task, such as~\cite{moya2018convolutional}, were not used as baselines because they were not designed for multi-step predictions over time series and, as such, would require significant changes.

\textbf{FFNN \cite{xiao2011hr_multistep}} is a feed-forward recursive architecture with skip connections using data from a wrist-worn 3-axial accelerometer. The model uses the average measurement of each sensor in a non-overlapping window of 30s. As the architecture details were not reported, we set the layer size to 16 and ReLU as the activation function for each layer based on random search optimization. We adapted their architecture, using a time window of $\tau_{\text{TS}}=4$s, to make it more comparable to our method (testing with $\tau_{\text{TS}} = 30$s as in the original work yielded worse results). We train the network using Adam, instead of the genetic algorithms used in that study.


\textbf{DeepConvLSTM \cite{ordonez2016convlstm}} performs a sequence of convolutions on the input series. The deep features from each time entry feed a LSTM. From each LSTM hidden vector, a prediction is computed using a single linear layer. In our adapted version, we add padding to the convolutions to maintain the length of the tensor and  select only the outputs corresponding to the last time input of each time snippet, as we have one label per time snippet. As in~\cite{ordonez2016convlstm}, we downsampled the input signals to 30Hz, and set time snippets' length to $\tau_{\text{TS}} = 3s$.


\section{Empirical evaluation} \label{sec:evaluation}

In this section we describe the experiments conducted to evaluate PCE-LSTM and their results. We begin with an overview of the experimental setup. Next, we present the IMU-based HR prediction experiments (the primary prediction task) and an ablation study on the PCE's impact on performance. Last, we describe the PPG-based HR estimation (the secondary prediction task).


\subsection{Experimental setup}

We evaluate the IMU-based multi-step HR estimation results w.r.t.\ Mean Absolute Error (MAE), Root Mean Squared Error (RMSE) \da{and Maximum Error ($\ell_1$)} as evaluation metrics. For the PPG-based HR estimation, however, we consider only MAE, since we transcribe the results from the paper where the baseline method was proposed~\cite{reiss2019deep}, which did not include another evaluation metric.

\noindent \textbf{Train-test split.}
We create several train-test splits using a ``leave one subject out strategy'', which
mimics a realistic setting where the model would be applied to individuals not contained in the training set, as is the case for most models trained offline~\cite{jordao2018human}. In a dataset with $S$ subjects, each of the subjects is used once as the test subject and the remaining $S-1$ subjects' time series are split into training segments of $N=50$ time snippets each and then randomly assigned to Train/Validation sets using an 80/20 split. For test, we use the whole series of each subject, which represents around one or two hours for most subjects in the PAMAP2 and DaLiA datasets respectively. With each of the $S$ subjects as the test subject, we perform 7 executions, with different Train/Validation splits and different neural network weight initialization, as done in \cite{reiss2019deep}.

\noindent \textbf{Hyperparameter Tuning.}
In order to choose PCM-LSTM' and the optimizer's (Adam) hyperparameters, we applied a random search using the results on PAMAP2's subject 5 as reference, following \da{\cite{ordonez2016convlstm}}. The following hyperparameters were selected: PCE$_F = 64$, $I = 12$, $\tau_{TS} = 4$, $r_{TS} = 0.5$, $\text{TSE}_{\text{out}} = 128$, $\text{TSE}_F = 16$, $\text{LSTM}_H = 64$, $\text{PCE}_{\text{out}} = 64$, $\text{TSE}_{\text{dropout}} = 0.15$,  learning rate $= 0.005$, weight decay $= 0.00005$.

\noindent \textbf{Training setup.}
 All the models are optimized with Adam using the $\ell_1$ loss as the cost function associated with the HR predictions. In each epoch, we compute the validation loss. After training is complete, we load the model weights that yielded the lowest validation loss.  Training was done using a batch size of 64 over 100 epochs for PCE-LSTM and DeepConvLSTM. FFNN showed slower convergence and hence was trained for 200 epochs. Subject 9 of the PAMAP2 dataset was not included in the analysis because the corresponding time series is shorter than the length of the training segment (102s).
 
\noindent \textbf{Mean vs.\ Ensemble performance.} Each method is trained 7 times for each test subject (using different train-validation splits). We compute a ``mean'' performance by averaging the errors of individual models and an ``ensemble'' performance by averaging the predictions and then computing the resulting error.


\subsection{IMU-based multi-step HR estimation}

\begin{table*}[t]
\centering
\setlength{\tabcolsep}{3.5pt}
\resizebox{\textwidth}{!}{
\small
\begin{tabular}{lrrrrrrrrrrrrrrr|r|r|r}
\toprule
\multirow{2}{*}{\textbf{Model}} &       \multicolumn{16}{c|}{\textbf{MAE by Test Subject [beats/minute]}} &  \multicolumn{1}{c}{\scriptsize RMSE}&  \multicolumn{1}{|c}{\scriptsize $\ell_1$}\\
{} &      \multicolumn{1}{c}{\textit{1}} &     \multicolumn{1}{c}{\textit{2}} &      \multicolumn{1}{c}{\textit{3}} &      \multicolumn{1}{c}{\textit{4}} &      \multicolumn{1}{c}{\textit{5}} &      \multicolumn{1}{c}{\textit{6}} &      \multicolumn{1}{c}{\textit{7}} &      \multicolumn{1}{c}{\textit{8}} &      \multicolumn{1}{c}{\textit{9}} &     \multicolumn{1}{c}{\textit{10}} &     \multicolumn{1}{c}{\textit{11}} &     \multicolumn{1}{c}{\textit{12}} &     \multicolumn{1}{c}{\textit{13}} &     \multicolumn{1}{c}{\textit{14}} &     \multicolumn{1}{c}{\textit{15}} & \multicolumn{1}{|c|}{\textit{Avg.}} & \multicolumn{1}{c|}{\textit{Avg.}} & \multicolumn{1}{c}{\textit{Avg.}}\\
\midrule 
FFNN (mean) &  13.2 &  10.5 &  12.9 &   9.0 &  45.0 &  \B32.1 &  13.7 &  16.0 &  \B12.3 &  \B11.9 &  \B22.2 &  17.8 &  17.2 &  12.7 &  14.0 &    17.4 & 21.8 & 64 \\
DeepConvLSTM (mean) &   9.8 &   7.7 &  16.1 &  12.9 &  44.1 &  34.9 &  15.8 &   \B9.2 &  14.8 &  13.2 &  25.5 &  \B11.3 &  15.3 &  13.6 &  11.2 &    17.0 & 22.2 & 69\\
PCE-LSTM (mean) &   \B9.3 &   \B6.5 &  \B12.2 &   \B8.7 &  \B42.0 &  34.7 &  \B11.3 &  12.2 &  13.6 &  12.2 &  22.4 &  15.1 &  \B10.9 &  \B11.5 &   \B9.2 &    \B15.5 & \B18.5 & \B51\\
\midrule 
FFNN (ens.) &  12.2 &  8.0 &  11.4 &   7.7 &  45.0 &  \B31.4 &  12.4 &  14.5 &  11.5 &   8.6 &  22.1 &  17.3 &  15.0 &  11.2 &  10.6 &    15.9 & 19.5 & 60\\
DeepConvLSTM (ens.) &   9.2 &  7.1 &  15.3 &  12.7 &  43.9 &  34.7 &  15.4 &   \B6.9 &  13.9 &  12.9 &  25.4 & \B10.7 &  15.1 &  13.2 &  10.5 &    16.5  & 20.6 & 68 \\
PCE-LSTM (ens.) &   \B8.4 &  \B5.1 &   \B7.8 &  \B 6.6 &  \B41.9 &  34.4 &   \B7.4 &   8.9 &  \B11.4 &   \B8.4 &  \B19.6 &  14.9 &   \B9.3 &   \B9.8 &   \B8.5 &    \B13.5 & \B16.2 & \B49 \\
\bottomrule
    \end{tabular}
}
\caption{Error Metrics on the DaLiA dataset (best shown in bold)}
\label{tab:benchmark_metrics_DaLiA}
\end{table*}

We begin by analyzing the performance metrics. Table~\ref{tab:benchmark_metrics_DaLiA} and \ref{tab:benchmark_metrics} shows the Mean and the Ensemble performances w.r.t. MAE by test subject and MAE averages for PCE-LSTM and the baselines, on DaLiA and PAMAP2 respectively. Due to space limitations, we only report \da{$\ell_1$ and} RMSE averages\fm{, but the complete results can be found in our technical report~\cite{deaguiar2021i}}. We note that the series of subjects 5 and 6 of DaLiA dataset can be regarded as outliers as none of the methods performed well on them. As expected, ensembles tend to outperform their standalone counterparts. Considering the ensemble performances, out of 23 subjects, FFNN, DeepConvLSTM and PCE-LSTM achieve the lowest errors for 2, 4 and 17 subjects, respectively. The lowest average error is obtained by the PCE-LSTM ensemble, with over 11.5\% lower MAE and 7.3\% lower RMSE than the next best method. \da{In relation to the maximum error, our method was over 18\% better than the next best method in the DALIA dataset, but just slightly better than the second best on the PAMAP2 dataset.} 

\begin{figure}[t]
    \centering
    \subfloat[\centering DALIA
    ]{{\includegraphics[width=.48\textwidth]{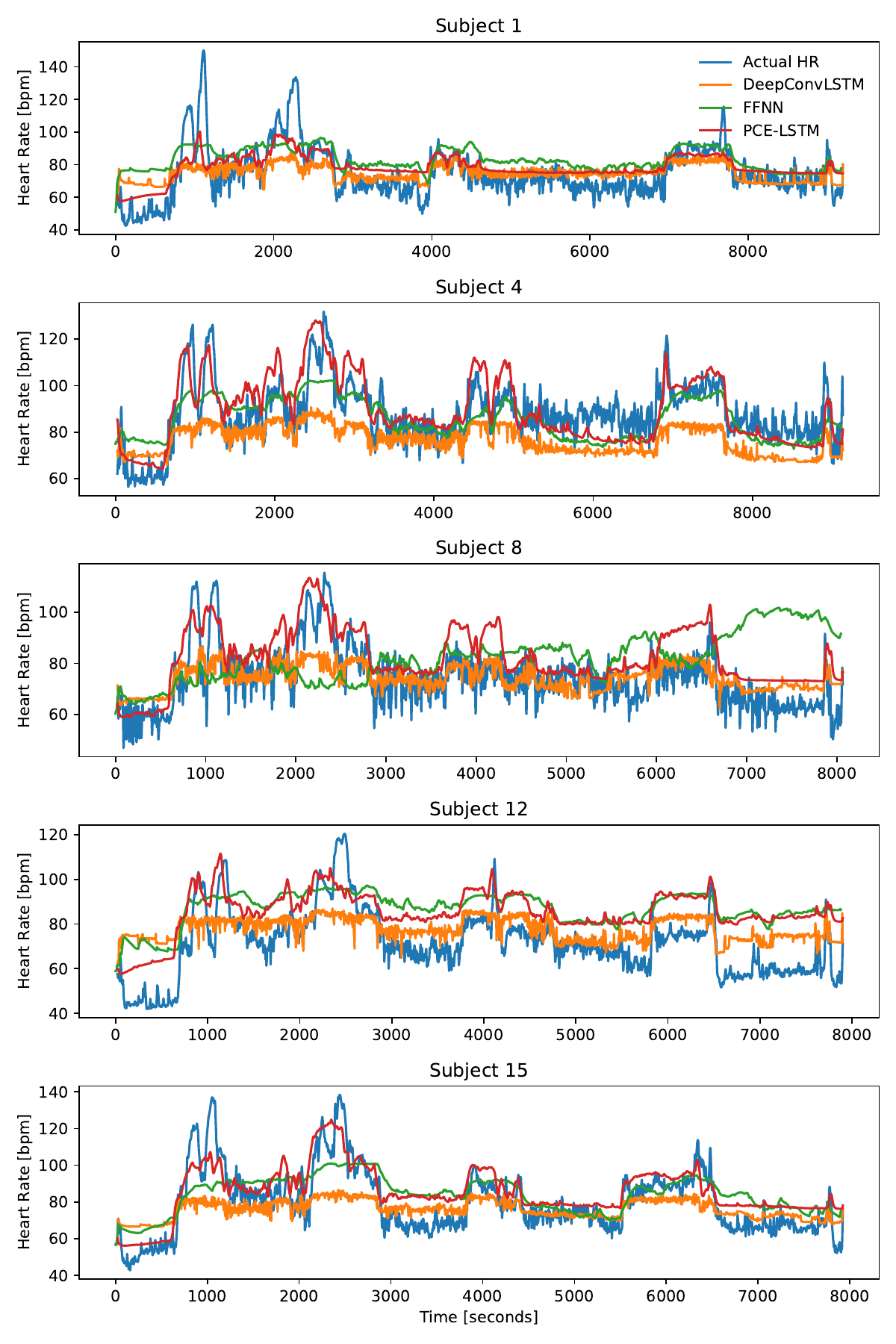} }}%
    \subfloat[\centering PAMAP2]{{\includegraphics[width=.48\textwidth]{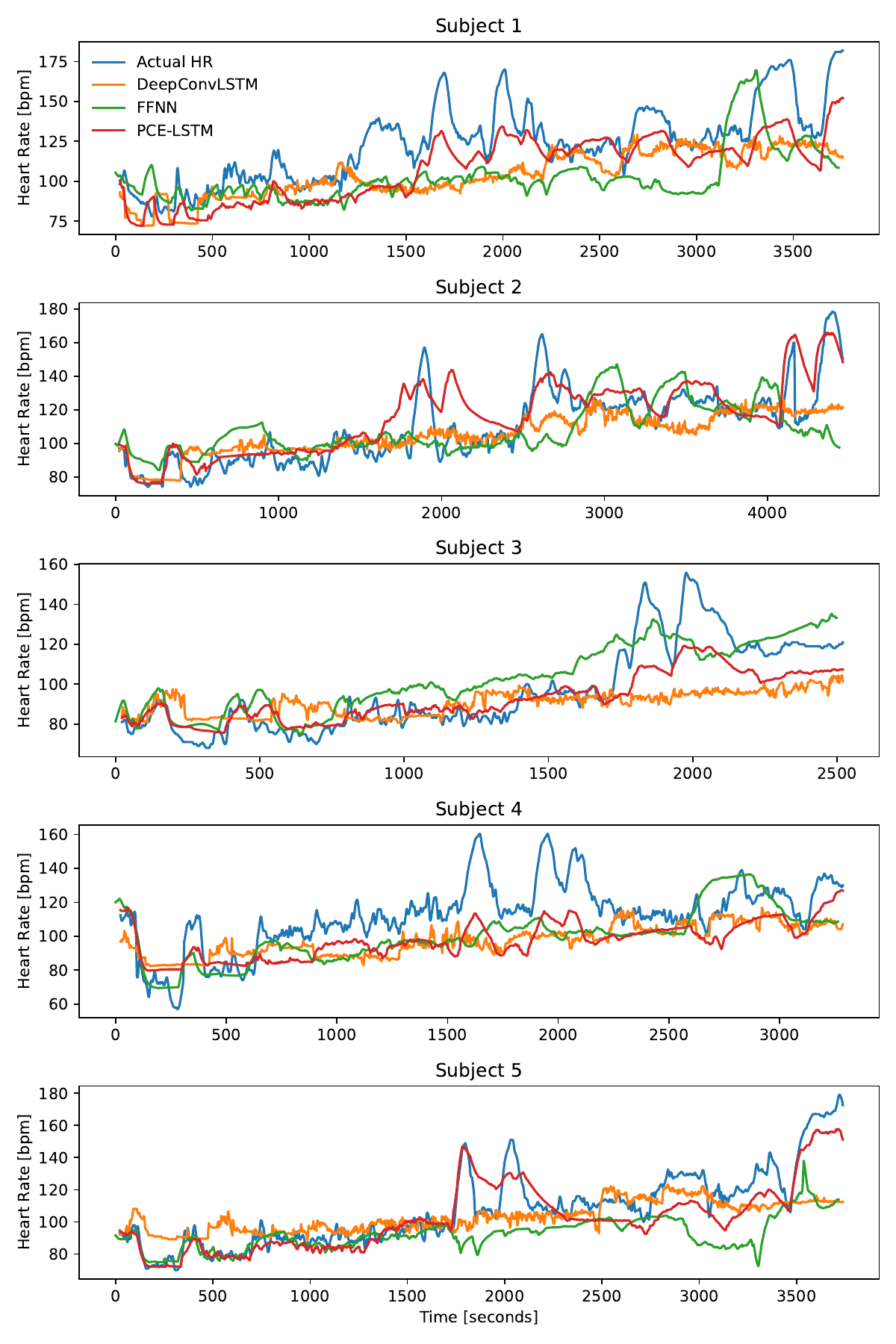} }}%
    \caption{Long-term IMU-based HR estimation}%
    \label{fig:samples}%
\end{figure}



\begin{table*}[t]
\centering
\resizebox{\textwidth}{!}{
\small
\setlength{\tabcolsep}{2.5pt}
\begin{tabular}{lrrrrrrrr|r|r|r|rrrrrrrr|r|r|r}
\toprule
 & \multicolumn{9}{c|}{\textbf{MAE by Test Subject [bpm]} (mean) } & \multicolumn{1}{c|}{\scriptsize RMSE} &
 \multicolumn{1}{c|}{\scriptsize $\ell_1$} & \multicolumn{9}{c|}{\textbf{MAE by Test Subject [bpm]} (ensemble)} & \multicolumn{1}{c}{\scriptsize RMSE}&
 \multicolumn{1}{|c}{\scriptsize $\ell_1$} \\
\textbf{Model} &       \multicolumn{1}{c}{\textit{1}} &       \multicolumn{1}{c}{\textit{2}} &       \multicolumn{1}{c}{\textit{3}} &       \multicolumn{1}{c}{\textit{4}} &       \multicolumn{1}{c}{\textit{5}} &       \multicolumn{1}{c}{\textit{6}} &       \multicolumn{1}{c}{\textit{7}} &       \multicolumn{1}{c|}{\textit{8}} & \multicolumn{1}{c|}{\textit{Avg.}} &  \multicolumn{1}{c|}{\textit{Avg.}} &
\multicolumn{1}{c|}{\textit{Avg.}} &
\multicolumn{1}{c}{\textit{1}} &       \multicolumn{1}{c}{\textit{2}} &       \multicolumn{1}{c}{\textit{3}} &       \multicolumn{1}{c}{\textit{4}} &       \multicolumn{1}{c}{\textit{5}} &       \multicolumn{1}{c}{\textit{6}} &       \multicolumn{1}{c}{\textit{7}} &       \multicolumn{1}{c|}{\textit{8}} & \multicolumn{1}{c|}{\textit{Avg.}} & \multicolumn{1}{c|}{\textit{Avg.}} & \multicolumn{1}{c}{\textit{Avg.}}\\
\midrule
FFNN                      &  24.9  &  18.3  &  10.5  &  \B17.2  &  18.6  &  14.7  &  21.6  &  33.7 & 19.9 & 26.1 & 72 & 23.2    &  14.2  &  9.5  &  \B15.8  &  17.8  &  13.4  &  20.3  &  23.9 & 17.3 & 22.1 & 64 \\
DeepConvLSTM              &  \B14.7  &  11.4  &   \B8.8  &  18.7  &  11.9  &  13.9  &  20.0  &  14.9 & 14.3 & 17.9 & 52 & \B11.8  &  10.6  &  \B8.3  &  18.4  &   9.3  &  13.1  &  19.5  &  13.3 & 13.0 & 15.8 & 46\\ 
PCE-LSTM &  16.6  &  \B10.7  &   9.0  &  19.5  &   \B9.6  &   \B9.9  &  \B12.6  &  \B14.8  & \B12.9  & \B16.4 & \B49& 16.5 &   \B9.5  &  8.4  &  16.7  &   \B8.4  &   \B8.8  &  \B10.8  &  \B13.2 & \B11.5 & \B14.6 & \B45  \\
\bottomrule
\end{tabular}
}
\caption{Error Metrics on PAMAP2 (best shown in bold)}
\label{tab:benchmark_metrics}
\end{table*}

For a qualitative evaluation, we plot the predictions of each model. Figure~\ref{fig:samples} shows the ensemble predictions for the complete series of five representative test subjects in the DaLiA and PAMAP2 datasets. We observe that DeepConvLSTM fails to capture the variance of the HR series: although the predictions are correlated with changes in HR, they tend to remain close to an ``average'' HR. FFNN, in turn, exhibits more variance, but cannot accurately capture the amplitude of the peaks and sometimes overestimates the HR. In contrast, PCE-LSTM (our method) can capture both peaks and valleys more accurately than the baselines.

\subsection{Performance Impact of PCE-LSTM's hidden state initialization}

In this section we conduct an ablation study to demonstrate that PCE-LSTM's strategy for initializing hidden state vectors is key to boosting the model's performance. For conciseness, we limit this study to the DaLiA dataset, which contains nearly twice the number of subjects, and a longer series per subject. A typical strategy to extract a hidden state is to use the LSTM itself to encode the relationship between sensor data and HR. In contrast, our model initializes the LSTM's hidden state vectors by passing the sensor and HR data corresponding to the first $I=12$ time snippets (26s) of the series through the Physical Conditioning Encoder (PC Encoder). As we assume that the PCE partially encodes an individual's physical conditioning, we expect that the joint training of the PCE-LSTM regression and the Discriminator will improve the regression predictions by promoting a better training of the PC Encoder subnetwork.

To quantify the performance impact of the proposed initialization and of using the discriminator, we conduct additional experiments with alternative initialization strategies. In what follows, ``with discr.'' indicates the joint training with the discriminator, ``without discr.'' indicates that the discriminator is not used; and ``LSTM self-encode'', indicates that hidden state vectors are initialized by feeding the HR to the network as an additional input channel in the TS of the Initialization Segment (replaced by zeros during the prediction segment).

\begin{table*}[t]
\setlength{\tabcolsep}{3pt}
\centering
\small
\resizebox{\textwidth}{!}{
\begin{tabular}{lrrrrrrrrrrrrrrr|r|r|r}
\toprule
{} &       \multicolumn{16}{c|}{\textbf{MAE by Test Subject [beats/minute]}} &  \multicolumn{1}{c}{\scriptsize RMSE} &  \multicolumn{1}{c}{\scriptsize $\ell_1$} \\
{} &      \multicolumn{1}{c}{\textit{1}} &     \multicolumn{1}{c}{\textit{2}} &      \multicolumn{1}{c}{\textit{3}} &      \multicolumn{1}{c}{\textit{4}} &      \multicolumn{1}{c}{\textit{5}} &      \multicolumn{1}{c}{\textit{6}} &      \multicolumn{1}{c}{\textit{7}} &      \multicolumn{1}{c}{\textit{8}} &      \multicolumn{1}{c}{\textit{9}} &     \multicolumn{1}{c}{\textit{10}} &     \multicolumn{1}{c}{\textit{11}} &     \multicolumn{1}{c}{\textit{12}} &     \multicolumn{1}{c}{\textit{13}} &     \multicolumn{1}{c}{\textit{14}} &     \multicolumn{1}{c}{\textit{15}} & \multicolumn{1}{|c|}{\textit{Avg.}}  & \multicolumn{1}{|c}{\textit{Avg.}} & \multicolumn{1}{|c}{\textit{Avg.}}\\
\midrule
LSTM self-encode (ens.)               &  11.0 &  7.6 &   \B7.1 &   7.4 &  48.6 &  33.6 &  10.7 &  15.6 &  11.5 &   9.3 &  21.2 &  15.2 &  10.7 &  10.7 &  10.4 &    15.4 & 18.3 & \B44\\
without discr. (ens.) &  10.5 &  5.9 &  13.7 &   7.6 &  43.7 &  \B29.5 &   7.8 &  14.3 &   \B9.4 &   9.5 &  \B18.8 &  \B13.6 &   \B9.1 &  11.4 &  \B7.7 &    14.2 & 16.8 & 46 \\
with discr. (ens.) &   \B8.4 &  \B5.1 &   7.8 &   \B6.6 &  \B41.9 &  34.4 &   \B7.4 &   \B8.9 &  11.4 &   \B8.4 &  19.6 &  14.9 &   9.3 &   \B9.8 &   8.5 &    \B13.5 & \B16.2 & 45\\
\bottomrule
\end{tabular}
}
\caption{Ablation: error results for variations of the PCE-LSTM model on DaLiA}
\label{tab:OurConvLSTM_varied_configurations_DaLiA}
\end{table*}

Table~\ref{tab:OurConvLSTM_varied_configurations_DaLiA} shows the prediction results for these experiments on DaLiA. PCE-LSTM's outperforms an LSTM self-encode initialization strategy even without a discriminator, but using the discriminator further reduces the error.

\subsection{PPG-based HR Estimation}

We adapt PCE-LSTM for PPG-based HR estimation. While similar to the IMU-based HR estimation, the former task has been better explored in the literature and has, therefore, well established baselines. For this reason, we compare PCE-LSTM to the SOTA method based on deep learning~\cite{reiss2019deep}. This method is based on CNNs and will be referred simply as CNN.   

To match the length and step size of each time snippet in~\cite{reiss2019deep}, we set $\tau_{\text{TS}} = 8$ and $r_{\text{TS}} = 0.75$ for this task.
Table~\ref{tab:ppg_metrics} shows the ensemble performance for each method on PPG-DaLiA and WESAD, when the individual designated in the column is the test subject. The results of CNN  were transcribed directly from \cite{reiss2019deep}. The last column shows the row average. On both datasets, PCE-LSTM provides error reductions of approximately 32\% when compared to the SOTA method. Another advantage of PCE-LSTM is that it has 2 orders of magnitude less parameters than the CNN -- approx.\ 120k parameters for PCE-LSTM vs.\ approx.\ 8.5M parameters for the CNN model.  

\begin{table*}[t]
\setlength{\tabcolsep}{3pt}
\centering
\small
\resizebox{\textwidth}{!}{
\begin{tabular}{ l c c c c c c c c c c c c c c c |c}
  \toprule 
   &  \multicolumn{16}{c}{\textbf{PPG-DaLiA: MAE by Test Subject [beats/minute]}}\\
 \cmidrule{2-17}
   \textbf{Model} & \textit{1} & \textit{2} & \textit{3}  & \textit{4} & \textit{5} & \textit{6} & \textit{7} & \textit{8} & \textit{9} & \textit{10} & \textit{11} & \textit{12} & \textit{13} & \textit{14} & \textit{15} & \multicolumn{1}{c}{\textit{Avg.}} \\  
  \cmidrule{2-17}
 \cite{reiss2019deep} & 7.73 & 6.74 & 4.03 &  5.90 &  18.51 & 12.88 &  3.91 & 10.87 & 8.79 &  4.03 &  9.22 &  9.35 &  4.29 &  4.37 &  4.17 & 7.65\\
  PCE-LSTM  & \B5.53 & \B3.77 & \B2.54 & \B5.41 & \B10.96 & \B5.54 & \B2.61 & \B9.09 & \B6.57 & \B2.62 & \B5.47 & \B8.47 & \B2.73 & \B3.71 & \B3.36 & \B5.22 \\
     \midrule
  &  \multicolumn{16}{c}{\textbf{WESAD: MAE by Test Subject [beats/minute]}}\\
   \cmidrule{2-17}
    \textbf{Model} & \textit{2} & \textit{3}  & \textit{4} & \textit{5} & \textit{6} & \textit{7} & \textit{8} & \textit{9} & \textit{10} & \textit{11} & \textit{13} & \textit{14} & \textit{15} & \textit{16} & \textit{17} & \multicolumn{1}{c}{\textit{Avg.}} \\

  \cmidrule{2-17}
 \cite{reiss2019deep} & 5.07 & 14.48 &  7.84 &  7.70 &  3.88 &  6.78 &  \B4.27 &  3.99 &  8.89 &  11.07 &  6.52 &  5.26 &  4.18 &  12.78 &  9.36 & 7.47 \\
 PCE-LSTM & \B3.59 & \B9.83 & \B3.46 & \B4.65 & \B2.65 & \B4.58 & 4.61 & \B3.03 & \B4.91 & \B7.06 & \B4.77 & \B4.67 & \B3.51 & \B4.91 & \B8.35 & \B4.97 \\
 \bottomrule
\end{tabular}
}
\caption{PPG-based HR estimation experiments } 
\label{tab:ppg_metrics}
\end{table*}

\section{Related Work}\label{sec:related}

Few studies used IMU sensors to predict the heart rate. \cite{yuchi2008heart} used a simple Feed-Forward Neural Network to predict the HR one step ahead, given its value and the average signal of each IMU sensor on the previous step. Quite similarly, \cite{xiao2011hr_multistep} performed a multi-step HR prediction by repeatedly using the HR computed for step $t$ to predict the HR for step $t+1$. Their experiments demonstrated promising results, but had some notable deficiencies: they used data from a single individual in his daily activities, hence without much variation in HR values.

Also on multi-step HR estimation, but using speed and acceleration as inputs, \cite{zhang2018multihr_bayers} proposed a Bayesian combined predictor, where one estimator was a linear regression and the other, a neural network similar to that of \cite{xiao2011hr_multistep}. In their experiments, data from multiple individuals performing running sessions was used, thus addressing some of the shortcomings in \cite{xiao2011hr_multistep}. However, the proposed method required calibration with actual HR data every 90s, hindering its practical use. 

For HAR, a task where the use of IMU data is widespread, \cite{hammerla2016comp_har} studied the performance of Deep Feed-Forward Networks (DNNs), CNNs and LSTMs, demonstrating that CNN and LSTM based models outperform DNNs and, among them, the best performing model was very dependent on the dataset used.

The hybrid CNN and LSTM model proposed by \cite{ordonez2016convlstm}, achieved good results but was surpassed by the CNN-based architecture of \cite{moya2018convolutional}.
More recently, some works proposed the use of self-attention based architectures for this task \cite{vaswani2017attention,wu2020deep_attention}. The model proposed by \cite{moya2018convolutional} outperformed the one proposed by \cite{vaswani2017attention} in the datasets that were common to both works. \cite{wu2020deep_attention} proposed a deep attention model that was able to surpass \cite{moya2018convolutional} in some, but not all datasets. A more extensive survey was conducted by \cite{wang2019survey}, where the authors note the variety of neural network architectures proposed for this task, such as CNNs, DNNs, RNNs, 
Stacked Auto Encoders and hybrid approaches.

A few studies have proposed methods for initializing state vectors of RNNs. To regularize and improve the dynamics of an RNN, \cite{rnn_init_bp:2006} proposed initializing the RNN with a vector proportional to the backpropagated errors. \cite{wenke2019contextual} showed that using contextual information to initialize the RNN hidden state improved the convergence rate, but not the final accuracy of the networks on some constructed problems. In modeling an aerial vehicle dynamics, \cite{rnn_init_ds} showed that passing an initial segment of the data through a custom network to set the hidden state vectors reduces errors, especially in the first iterations of predictions. Our work differs from theirs in several ways. Most markedly, (i) it uses a discriminator to improve the ability of the network to capture the subjects' inherent characteristics based on previous activities and (ii) it is able to improve predictions over windows of up to 2 hours. This corroborates the idea that PCE-LSTM is encoding physical conditioning, rather than information about specific physical activities.


\section{Conclusions}\label{sec:impact}

We investigated the much neglected task of predicting HR from IMU sensor data. We started from the premise that, depending on their physical conditioning, different people will display different HR levels when performing the same exercises. We proposed a neural architecture dubbed Physical Conditioning Embedding LSTM (PCE-LSTM) that employs a convolutional network to extract vectors which carry information about the relationship between sensor measurements and the HR for a specific individual, thus representing his/her physical conditioning. These vectors are used as the initial state vectors for a LSTM network that outputs HR predictions from sensor data. We evaluate the prediction accuracy of PCE-LSTM w.r.t.\ MAE and RMSE using public datasets (PAMAP2, PPG-DaLiA, WESAD). Moreover, we compare PCE-LSTM with 2 baselines: one model proposed for this task (FFNN) and one minimally-adapted state-of-the-art model (DeepConvLSTM) originally proposed for the closely related task known as Human Activity Recognition.

PCE-LSTM yields over 10\% (resp.\ 30\%) lower MAE in the IMU-based (resp.\ PPG-based) HR estimation task. Last, we conduct additional experiments to show that the performance gains achieved by our method are, in part, due to the strategy used to initialize the hidden vectors. Specifically, using the outputs of the PCE network applied to the data from the previous 12 time snippets of the subject's time series is helpful and works better than using the LSTM itself to output hidden vectors from the same data, especially when using the PCE discriminator during training.

We have highlighted the potential of neural network approaches to tackle this problem. Some questions are left open for future work, such as whether these results are accurate enough to be successfully used in real-world applications or whether the metrics used best represent the suitability of a model. We hope that this work can foster the development of new methods and metrics for this task, bringing about a practical use for the method in medical or fitness applications.


%
\bibliographystyle{unsrt} 
\bibliography{references}
\end{document}